\documentclass[3p,times,procedia]{elsarticle}
\usepackage{ecrc}
\usepackage[bookmarks=false]{hyperref}
    \hypersetup{colorlinks,
      linkcolor=blue,
      citecolor=blue,
      urlcolor=blue}

\volume{00}

\firstpage{1}

\journalname{Procedia Computer Science}

\runauth{Fanny and Cenggoro}

\jid{procs}

\usepackage{amssymb}

\usepackage[figuresright]{rotating}

\usepackage{multirow}
\usepackage{subfig}
\usepackage{graphicx}

\begin{document}
\begin{frontmatter}




\title{Deep Learning for Imbalance Data Classification using Class Expert Generative Adversarial Network}


\author[a]{Fanny} 
\author[a,b]{Tjeng Wawan Cenggoro}

\address[a]{Computer Science Department, School of Computer Science, Bina Nusantara University, Jakarta, Indonesia 11480}
\address[b]{Bioinformatics and Data Science Research Center, Bina Nusantara University, Jakarta, Indonesia 11480}

\begin{abstract}
Without any specific way for imbalance data classification, artificial intelligence algorithm cannot recognize data from minority classes easily. In general, modifying the existing algorithm by assuming that the training data is imbalanced, is the only way to handle imbalance data. However, for a normal data handling, this way mostly produces a deficient result. In this research, we propose a class expert generative adversarial network (CE-GAN) as the solution for imbalance data classification. CE-GAN is a modification in deep learning algorithm architecture that does not have an assumption that the training data is imbalance data. Moreover, CE-GAN is designed to identify more detail about the character of each class before classification step. CE-GAN has been proved in this research to give a good performance for imbalance data classification.
\end{abstract}

\begin{keyword}
imbalance data; classification; deep learning; generative adversarial network




\end{keyword}
\end{frontmatter}




\section{Introduction}
\label{introduction}

Classification of imbalance data is one of a classic problem in the artificial intelligence area, especially for classification in machine learning. Imbalance data has been proved can decrease the performance of machine learning algorithm \cite{Dalyac2014}, where imbalance data means the total of data from each class is significantly different.
The example of imbalance data can be seen in the works from Cenggoro et al. \cite{Cenggoro2016, Cenggoro2017}. The data from those researches shows that classes of an urban area is up to 62.45\% from the total area, while the class of an open area is only 0.02\% from the total area. Usually, machine learning algorithm cannot learn this kind of data, especially for minority classes because the information is covered by majority classes data.
There are some researches in deep learning area that is trying to prevent classification in imbalance data \cite{Huang2016,Khan2017}. However, these researches still use algorithm that assumes the training dataset is imbalance. This assumption can decrease the performance of the algorithm if it is used for balance training dataset.

\section{Related Works}
\label{related}
Several works have been done to solve imbalance data problem in deep learning. Huang et al. \cite{Huang2016} handled imbalance data by integrating sampling process into Convolutional Neural Network (CNN) \cite{LeCun1989}. The sampling process was done by dividing the data into several clusters, then a pair of data based on the determined condition was retrieved. These pairs of data were then inputted to the CNN. After this, the process was then repeated again from clustering until the model reaches the convergences.
Yan et al. \cite{Yan2015} and Khan et al. \cite{Khan2017} handled imbalance data by modifying a CNN technique. Yan et al. modified it by combining bootstrapping and CNN while Khan et al. designed a cost function that is tailored to CNN characteristics to solve the same problem.
In addition, there are several researches that explained how to handle imbalance data, even though their main focus was not specifically to deal with imbalance data. Wei et al. \cite{Wei2015} handled an imbalance data in object recognition by designing an image-cutting technique that is used as an additional training data. George \cite{George2015} and Mostajabi et al. \cite{Mostajabi2015} used cost-sensitive to handle imbalance data in scene-parsing. The technique that was used in these researches is designed only for the dataset that used in each research. Thus, it would be difficult to implement the technique for different dataset.

\section{Proposed Method}
\label{method}

\subsection{Class Experts}
Class Experts (CE) is a concept proposed by Cenggoro  to overcome imbalance problem in deep learning \cite{Cenggoro2016a}. This concept utilizes a paradigm of arranging neural network layers in parallel, which each layer has been pretrained to recognize characteristic of a single class in the training data. These layers are pretrained with autoencoder-style algorithm. Cenggoro specifically use Variational Autoencoder in his work \cite{Cenggoro2016a}. After the pretraining, the weights are transferred to main neural network model and trained supervised like a normal classification task. Fig. \ref{fig:ce} depicted the main model with parallel layers grouped as Class Experts Layer, which each module in this layer is pretrained using single-class dataset.

\subsection{Class Experts Generative Adversarial Network}
\begin{figure}[ht!]
\parbox[t][][t]{.48\linewidth}{
\centering
\includegraphics[width=0.48\textwidth]{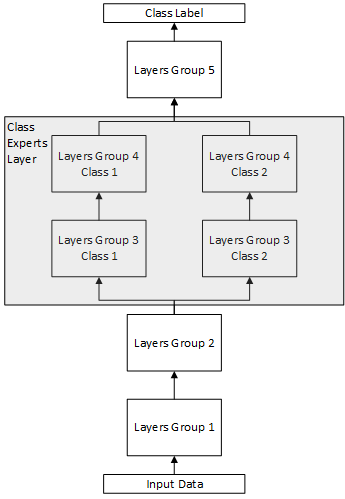}
\caption{Example of Main Neural Network Model with Class Experts Layer}
\label{fig:ce}
}
\parbox[t][][t]{.04\linewidth}{\hfill}
\parbox[t][][t]{.48\linewidth}{
\centering
\includegraphics[width=0.48\textwidth]{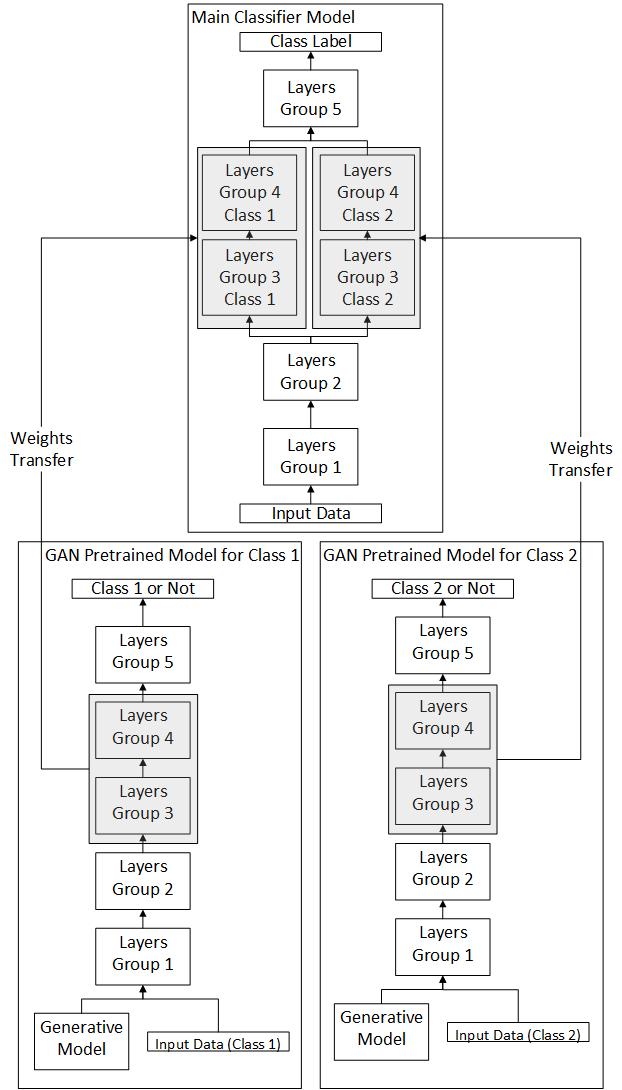}
\caption{Illustration of GAN Pretrained Weights Transfer Process for Two Classes}
\label{fig:transfer}
}
\end{figure}

Different from the work of Cenggoro \cite{Cenggoro2016a}, we propose the use of Generative Adversarial Network (GAN) \cite{Goodfellow2014} for pretraining algorithm instead of autoencoder-style algorithm. In contrast to autoencoder-style algorithm, GAN has more theoretical benefit to encode single-class characteristics useful for classification task. The most noticeable benefit of GAN for classification task is that it incorporates the use of discriminative model in its process. When trained with only single class, the GAN discriminative model can be viewed as a classifier that is able to tell whether input data is from the assigned class or not. In other words, this discriminative model encode distribution $p(y_i |x)$, where $y$ is the decision whether the input data $x$ is from class $i$ or not. This is contrast to autoencoder-style algorithm that encodes distribution $p(x)$. Therefore, the weights of GAN discriminative model are more natural to be transferred to the main classifier model, which optimize distribution $p(y|x)$, where $y$ is the label to determine the class of input data $x$ in multi-class setting. Fig. \ref{fig:transfer} illustrates the process of GAN pretrained weights transfer to main CNN model in the proposed algorithm. After the weight transfer, we can choose whether to keep the weight frozen or not during the finetuning phase. To the rest of this paper, we will call the proposed algorithm with frozen weights as frozen class expert generative adversarial network (FCE-GAN). For the counterpart with trainable weights, it will be called as class expert generative adversarial network (CE-GAN).

\section{Experiment Setting}
\label{setting}

\subsection{Data Preparation}
\begin{figure}
\centering
\includegraphics[width=0.7\textwidth]{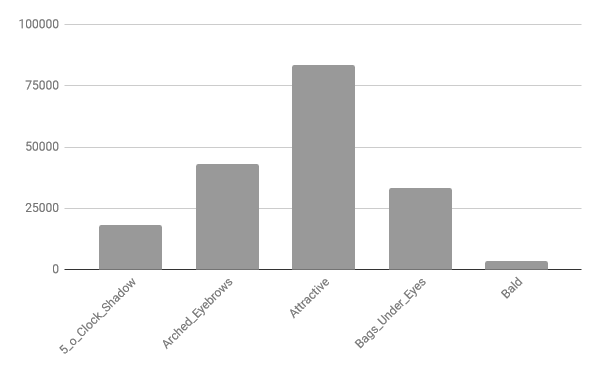}
\caption{Classes Distribution of Training Dataset}
\label{fig:chart}
\end{figure}

The data used in this research is sliced from CelebA dataset \cite{liu2015deep}. Specifically, only first five classes of CelebA dataset used in this research: 5 O'Clock Shadow, Arched Eyebrows, Attractive, Bags Under Eyes, ands Bald. The limitation of dataset used is necessary for the conciseness of balancing effect analysis in this research. The whole sliced dataset is then divided into training, validation, and testing sub-dataset, with the ratio of 6:2:2. The distribution of training sub-dataset is visualized in Fig. \ref{fig:chart}. From this figure, we can see that the training dataset is imbalance, with Attractive as the majority class and Bald as the most minority class.

\subsection{Models in Comparison}
In this research, performance of four models are compared: two proposed algorithms (CE-GAN CNN and FCE-GAN CNN) and two other algorithms from previous researches. The other two compared algorithm are Bootstrapping CNN \cite{Yan2015} and CosenCNN \cite{Khan2017}, which are CNN models designed to tackle imbalance problems. In addition, we also display the result from pure baseline CNN, which is explained in sub-chapter \ref{configuration}.

\subsection{Model Configuration}
\label{configuration}
The CNN used in all compared algorithms are based on a same CNN architecture, which is called as baseline CNN architecture in this paper. The objective of using this baseline CNN is to guarantee that the performance comparison is not affected by the different CNN capability used by each compared algorithm. The architecture used for the baseline CNN is taken from the discriminative CNN used in the GAN model, which is described in table \ref{tab:arch_disc}. Layer 1 to 6 of the discriminative architecture is based on the reverse of layer 1 to 6 of the generative CNN architecture employed in the GAN model with a little adjustment so that discriminative CNN has same kernel size for all its convolution layer. The generative CNN architecture is described in table \ref{tab:arch_gen}. The architecture of generative CNN is designed specifically to generate image with the size of 218x178, same as the original images size, from a randomly generated input vector with size of 1x100.

For a better performance, we employ several batch normalizations within the discriminative CNN architecture, noted as BN in table \ref{tab:arch_gen} and \ref{tab:arch_disc}. Batch normalization has been proved to increase deep learning model performance \cite{Ioffe2015}. We also use rectified linear unit (ReLU) activation functions for our CNN. ReLU has been proved to be able to boost the performance of CNN with deep layers \cite{Krizhevsky2012}.

The optimization method used for all compared models is Adam \cite{Kingma2015}, which is already proven to be able to optimize deep learning model faster. The learning rate used is 0.001, as suggested by Kingma \& Ba \cite{Kingma2015}.
To determine the number of consecutive layers used as class experts in the proposed models, a pre-experiment is run to see the validation accuracy of different number of layer in CE for CE GAN CNN in 5 \textsuperscript{th} iteration. The result of this pre-experiment is tabularized in table \ref{tab:preexp}. From this table, layer 3, 4, 5, and 6 give the best results when used in CE of the CE GAN CNN. Therefore, this CE architecture configuration is used for performance comparison of proposed models with other algorithms.

\begin{table}[h]
\parbox[t][][t]{.48\linewidth}{
\caption{Generator CNN Architecture}
\begin{tabular*}{\hsize}{@{\extracolsep{\fill}}lll@{}}
\label{tab:arch_gen}
\toprule
\#Layer & Layer Type & Layer Description \\
\colrule
1 & Deconvolution+BN+ReLU  & Size = 5x4x192 \\
 & & Stride = 1 \\
 & & Padding Zero = No \\
 & & Input Size = 1x100x1 \\
 & & Output Size = 5x4x192 \\
\colrule
2 & Deconvolution+BN+ReLU  & Size = 4x4x192 \\
 & & Stride = 5 \\
 & & Padding Zero = Yes \\
 & & Output Size = 25x20x192 \\
\colrule
3 & Deconvolution+BN+ReLU  & Size = 3x3x192 \\
 & & Stride = 1 \\
 & & Padding Zero = No \\
 & & Output Size = 27x22x192 \\
\colrule
4 & Deconvolution+BN+ReLU  & Size = 4x4x96 \\
 & & Stride = 2 \\
 & & Padding Zero = Yes \\
 & & Output Size = 54x44x96 \\
\colrule
5 & Deconvolution+BN+ReLU  & Size = 4x4x48 \\
 & & Stride = 2 \\
 & & Padding Zero = Yes \\
 & & Output Size = 108x88x48 \\
\colrule
6 & Deconvolution+BN+ReLU  & Size = 4x4x24 \\
 & & Stride = 2 \\
 & & Padding Zero = No \\
 & & Output Size = 218x178x24 \\
\colrule
7 & Deconvolution  & Size = 4x4x3 \\
 & & Stride = 1 \\
 & & Padding Zero = Yes \\
 & & Output Size = 218x178x3 \\
\botrule
\end{tabular*}
}
\parbox[t][][t]{.04\linewidth}{
\begin{tabular*}{\hsize}{@{\extracolsep{\fill}}lll@{}}
\end{tabular*}
}
\parbox[t][][t]{.48\linewidth}{
\caption{Discriminator CNN Architecture}
\begin{tabular*}{\hsize}{@{\extracolsep{\fill}}lll@{}}
\label{tab:arch_disc}
\toprule
\#Layer & Layer Type & Layer Description \\
\colrule
1 & Convolution+ReLU  & Size = 4x4x24 \\
 & & Stride = 2 \\
 & & Padding Zero = Yes \\
 & & Input Size = 218x178x3 \\
 & & Output Size = 109x89x24 \\
\colrule
2 & Convolution+BN+ReLU  & Size = 4x4x48 \\
 & & Stride = 2 \\
 & & Padding Zero = Yes \\
 & & Output Size = 55x45x48 \\
\colrule
3 & Convolution+BN+ReLU  & Size = 4x4x96 \\
 & & Stride = 2 \\
 & & Padding Zero = Yes \\
 & & Output Size = 28x23x96 \\
\colrule
4 & Convolution+BN+ReLU  & Size = 4x4x192 \\
 & & Stride = 2 \\
 & & Padding Zero = Yes \\
 & & Output Size = 14x12x192 \\
\colrule
5 & Convolution+BN+ReLU  & Size = 4x4x192 \\
 & & Stride = 2 \\
 & & Padding Zero = Yes \\
 & & Output Size = 7x6x192 \\
\colrule
6 & Convolution+BN+ReLU  & Size = 4x4x24 \\
 & & Stride = 2 \\
 & & Padding Zero = Yes \\
 & & Output Size = 4x3x192 \\
\colrule
7 & Convolution & Size = 4x4x3 \\
 & & Stride = 1 \\
 & & Padding Zero = No \\
 & & Output Size = 3x2x1 \\
\colrule
8 & Fully Connected + ReLU & \#Neuron = 200 \\
 & & Output Size = 200x1 \\
\colrule
9 & Fully Connected + Sigmoid & \#Neuron = 5 \\
 & & Output Size = 5x1 \\
\botrule
\end{tabular*}
}
\end{table}

The optimization method used for all compared models is Adam \cite{Kingma2015}, which is already proven to be able to optimize deep learning model faster. The learning rate used is 0.001, as suggested by Kingma \& Ba \cite{Kingma2015}.
To determine the number of consecutive layers used as class experts in the proposed models, a pre-experiment is run to see the validation accuracy of different number of layer in CE for CE-GAN CNN in 5th iteration. The result of this pre-experiment is tabularized in table \ref{tab:preexp}. From this table, layer 3, 4, 5, and 6 give the best results when used in CE of the CE-GAN CNN. Therefore, this CE architecture configuration is used for performance comparison of proposed models with other algorithms.

\begin{table}[h]
\caption{Pre-Experiment Result on Determining Number of Layers Used as CE.}
\begin{tabular*}{\hsize}{@{\extracolsep{\fill}}lll@{}}
\label{tab:preexp}
\toprule
Layers Used as CE & Validation Accuracy \\
\colrule
6 & 75.59 \%  \\
5 and 6 & 75.59 \%  \\
4, 5, and 6 & 74.92 \%  \\
\textbf{3, 4, 5, and 6} & \textbf{76.27 \%}  \\
2, 3, 4, 5, and 6 & 73.22 \%  \\
\botrule
\end{tabular*}
\end{table}

\subsection{Analyzing Results}
\label{analyzing}
The metrics used to assess performance of all compared algorithms are classification accuracy and precision. Classification accuracy used to assess overall algorithm performance. Equation \ref{eq:acc} shows the calculation of classification accuracy.

\begin{equation}
\label{eq:acc}
Accuracy = \frac{TP+TN}{TP+TN+FP+FN}
\end{equation}

Where:
\begin{itemize}[]
\item TP = True Positive, the number of data that are correctly classified as a particular class
\item TN = True Negative, the number of data that are correctly not classified as a particular class
\item FP = False Positive, the number of data that are mistakenly classified as a particular class
\item FN = False Negative, the number of data that are mistakenly  not classified as a particular class
\end{itemize}

To assess the balancing effect of all compared algorithms, we consider the use of classification precision metric. Classification precision can be calculated using equation \ref{eq:prec}. By using precision, we can see the performance of each algorithm in terms of correctly classify minority class (TP), without being shadowed by the sheer number of negative examples from minority class. 

\begin{equation}
\label{eq:prec}
Precision = \frac{TP}{TP+FP}
\end{equation}

\section{Experiment Results and Analysis}
\label{results}
\begin{figure}
     \centering
     \subfloat[][]{\includegraphics[width=0.3\textwidth]{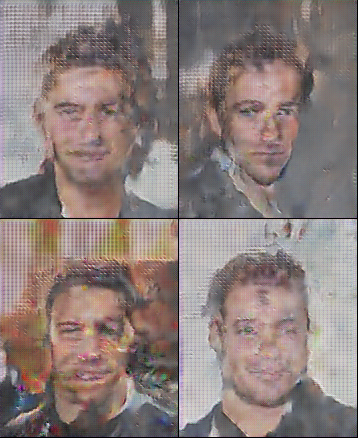}\label{fig:fce_0}} \hspace{1em}
     \subfloat[][]{\includegraphics[width=0.3\textwidth]{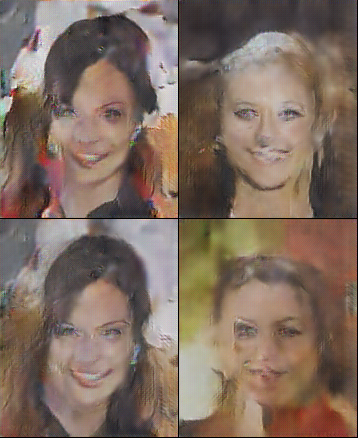}\label{fig:fce_1}} \hspace{1em}
     \subfloat[][]{\includegraphics[width=0.3\textwidth]{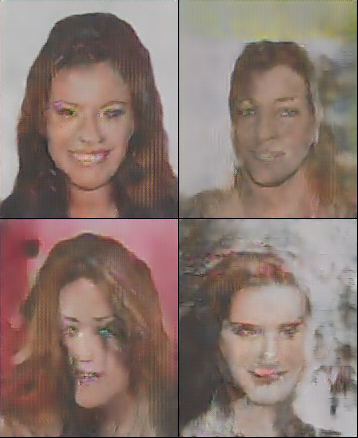}\label{fig:fce_2}}\\
     \subfloat[][]{\includegraphics[width=0.3\textwidth]{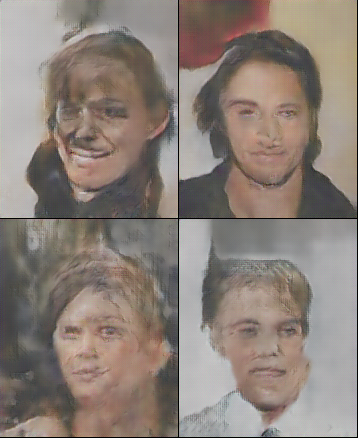}\label{fig:fce_3}} \hspace{1em}
     \subfloat[][]{\includegraphics[width=0.3\textwidth]{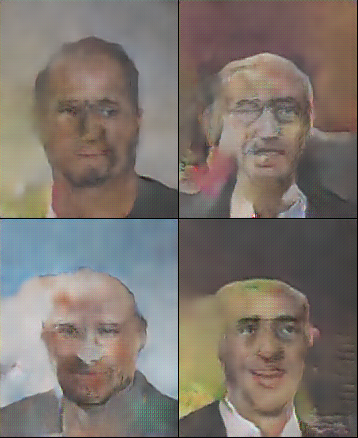}\label{fig:fce_4}}
     \caption{Sample of GAN Generated Images Pretrained with class: (a) 5 O'Clock Shadow; (b) Arched Eyebrows; (c) Attractive; (d) Bags Under Eyes; (e) Bald.}
     \label{fig:gen_img}
\end{figure}

The performance of all GAN pretrained models can bee seen by looking at the sample of generated images by the model in figure \ref{fig:gen_img}. We can see that the models generally able to catch unique features for each classes. For instance, all generated images in class 5 o'clock shadow, arched eyebrows, bags under eyes, and bald indeed has 5 o'clock shadow, arched eyebrow, bags under eyes, and baldness. While it would be difficult to see if model for subjective class such as attractive is able to catch desirable feature, we can still see that the model is able to generate reasonable face. It is important to make sure that the model can catch unique features of each class, as the idea of CE is to transfer these features to the classification model.

Table \ref{tab:accuracy} shows classification accuracy performance of all compared algorithms. We can see that the best overall accuracy is achieved by CE-GAN CNN. The best classification accuracy of each class also performed by CE-GAN CNN, with the only exception for class Bags Under Eyes, which only 0.45\% less than the best accuracy achieved by baseline CNN.

\begin{table}[h]
\caption{Compared Algorithms Classification Accuracy Performance.}
\begin{tabular*}{\hsize}{@{\extracolsep{\fill}}lllllll@{}}
\label{tab:accuracy}
\toprule
\multirow{2}{*}{Algorithm} & \multicolumn{5}{c}{Per-Class Accuracy} & \multirow{2}{*}{Overall Accuracy} \\
\cline{2-6} 
 & 5 O'Clock Shadow & Arched Eyebrows & Attractive & Bags Under Eyes & Bald & \\
\colrule
CNN with Bootstrapping & 84.85 & 67.14 & 71.26 & 73.80 & 95.97 & 78.61  \\
CosenCNN & 86.43 & 67.95 & 72.60 & 75.39 & 97.13 & 79.90  \\
Baseline CNN & 92.04 & 78.95 & 81.23 & \textbf{85.03} & 98.40 & 87.13  \\
CE-GAN CNN & \textbf{92.71} & \textbf{78.98} & \textbf{82.09} & 84.58 & \textbf{98.56} & \textbf{87.38}  \\
FCE-GAN CNN & 88.35 & 68.86 & 78.73 & 81.52 & 97.48 & 82.99  \\
\botrule
\end{tabular*}
\end{table}

As previously discussed, it is necessary to see the performance of all compared algorithms using classification precision metric to analyze their balancing effect. These precision results are tabularized in table \ref{tab:precision}. We can see from this table that CE-GAN CNN has the best balancing effect among all compared algorithms. This can be seen from the fact that only CE-GAN CNN can increase all the precision of minority classes from baseline CNN. Not only that, CE-GAN CNN also able to increase the most majority class precision by 1.57\%.
Meanwhile, the other approach proposed in this paper, FCE-GAN, is not performing as expected. While it only differs from CE-GAN CNN in keeping the transferred weights fixed, the precision performance of FCE-GAN CNN is significantly lower than CE-GAN CNN. Its balancing effect even lower than CNN with bootstrapping, which can increase two of four minority classes precision.

\begin{table}[h]
\caption{Compared Algorithms Classification Precision Performance.}
\begin{tabular*}{\hsize}{@{\extracolsep{\fill}}lllllll@{}}
\label{tab:precision}
\toprule
\multirow{2}{*}{Algorithm} & \multicolumn{5}{c}{Per-Class Accuracy} & \multirow{2}{*}{Overall Precision} \\
\cline{2-6} 
 & 5 O'Clock Shadow & Arched Eyebrows & Attractive & Bags Under Eyes & Bald & \\
\colrule
CNN with Bootstrapping & \textbf{82.30} & 38.89 & 71.05 & 8.53 & 69.74 & 52.54 \\
CosenCNN & 9.18 & 48.52 & \textbf{89.81} & 26.55 & 0.00 & 58.54 \\
BaseLine CNN & 50.10 & 73.44 & 84.08 & 73.40 & 69.50 & 76.02 \\
CE-GAN CNN & 65.20 & \textbf{76.00} & 85.65 & \textbf{77.35} & \textbf{77.78} & \textbf{79.64} \\
FCE-GAN CNN & 56.82 & 35.63 & 86.62 & 53.45 & 65.72 & 64.30 \\
\botrule
\end{tabular*}
\end{table}

\section{Conclusion}
In this paper, we demonstrate the benefits of using CE paradigm and GAN pretraining in countering imbalance data problem in CNN. The utilization of CE is able not only to give a better balancing precision, but also to increase the overall accuracy of standard CNN. The balancing effect of given by CE is also better than other compared algorithm in this paper. However, to achieve these benefits, the transferred weights from pretrained GAN model need to be allowed to change as the main CNN model learn supervisedly.

We might notice that the type of GAN employed in this research is the original GAN. There has been several type of GANs that modify the original GAN and succesfully improve its performance. Therefore, it would be interesting to see the effect of using CE concept with other type of GANs.

\section*{Acknowledgements}
This research is funded by Bina Nusantara University Applied Technology Grant 2017. We would also like to acknowledge NVIDIA – Bina Nusantara University AI R\&D Center for the GPU used in this research.





\biboptions{super}
\bibliographystyle{model3a-num-names}
\bibliography{procs-template-ICCSCI_2018}

\begin{thebibliography}{16}
\providecommand{\natexlab}[1]{#1}
\providecommand{\url}[1]{\texttt{#1}}
\providecommand{\urlprefix}{URL }
\expandafter\ifx\csname urlstyle\endcsname\relax
  \providecommand{\doi}[1]{doi:\discretionary{}{}{}#1}\else
  \providecommand{\doi}{doi:\discretionary{}{}{}\begingroup
  \urlstyle{rm}\Url}\fi
\providecommand{\eprint}[2][]{\url{#2}}
\providecommand{\bibinfo}[2]{#2}
\ifx\xfnm\undefined \def\xfnm[#1]{\unskip,\space#1}\fi
\makeatletter\def\@biblabel#1{#1.}\makeatother
\bibitem[{Dalyac et~al.(2014)Dalyac, Shanahan and Kelly}]{Dalyac2014}
\bibinfo{author}{Dalyac\xfnm[ A.]}, \bibinfo{author}{Shanahan\xfnm[ M.]},
  \bibinfo{author}{Kelly\xfnm[ J.]}.
\newblock \bibinfo{title}{{Tackling Class Imbalance with Deep Convolutional
  Neural Networks}}; \bibinfo{year}{2014}.
\bibitem[{Cenggoro et~al.(2016)Cenggoro, Isa and Kusuma}]{Cenggoro2016}
\bibinfo{author}{Cenggoro\xfnm[ T.W.]}, \bibinfo{author}{Isa\xfnm[ S.M.]},
  \bibinfo{author}{Kusuma\xfnm[ G.P.]}.
\newblock \bibinfo{title}{{Construction of Jakarta Land Use/Land Cover dataset
  using classification method}}.
\newblock In: \emph{\bibinfo{booktitle}{2016 IEEE Region 10 Symposium
  (TENSYMP)}}. \bibinfo{publisher}{IEEE}.
\newblock ISBN \bibinfo{isbn}{978-1-5090-0931-2}; \bibinfo{year}{2016},
  \hspace{0pt}p. \bibinfo{pages}{337--342}.
\newblock \doi{\bibinfo{doi}{10.1109/TENCONSpring.2016.7519429}}.
\bibitem[{Cenggoro et~al.(2017)Cenggoro, Isa, Kusuma and
  Pardamean}]{Cenggoro2017}
\bibinfo{author}{Cenggoro\xfnm[ T.W.]}, \bibinfo{author}{Isa\xfnm[ S.M.]},
  \bibinfo{author}{Kusuma\xfnm[ G.P.]}, \bibinfo{author}{Pardamean\xfnm[ B.]}.
\newblock \bibinfo{title}{{Classification of imbalanced land-use/land-cover
  data using variational semi-supervised learning}}.
\newblock In: \emph{\bibinfo{booktitle}{2017 International Conference on
  Innovative and Creative Information Technology (ICITech)}}.
  \bibinfo{publisher}{IEEE}.
\newblock ISBN \bibinfo{isbn}{978-1-5386-4046-3}; \bibinfo{year}{2017},
  \hspace{0pt}p. \bibinfo{pages}{1--6}.
\newblock \doi{\bibinfo{doi}{10.1109/INNOCIT.2017.8319149}}.
\bibitem[{Huang et~al.(2016)Huang, Li, Loy and Tang}]{Huang2016}
\bibinfo{author}{Huang\xfnm[ C.]}, \bibinfo{author}{Li\xfnm[ Y.]},
  \bibinfo{author}{Loy\xfnm[ C.C.]}, \bibinfo{author}{Tang\xfnm[ X.]}.
\newblock \bibinfo{title}{{Learning Deep Representation for Imbalanced
  Classification}}.
\newblock In: \emph{\bibinfo{booktitle}{2016 IEEE Conference on Computer Vision
  and Pattern Recognition (CVPR)}}. \bibinfo{publisher}{IEEE}.
\newblock ISBN \bibinfo{isbn}{978-1-4673-8851-1}; \bibinfo{year}{2016},
  \hspace{0pt}p. \bibinfo{pages}{5375--5384}.
\newblock \doi{\bibinfo{doi}{10.1109/CVPR.2016.580}}.
\bibitem[{Khan et~al.(2017)Khan, Hayat, Bennamoun, Sohel and
  Togneri}]{Khan2017}
\bibinfo{author}{Khan\xfnm[ S.H.]}, \bibinfo{author}{Hayat\xfnm[ M.]},
  \bibinfo{author}{Bennamoun\xfnm[ M.]}, \bibinfo{author}{Sohel\xfnm[ F.A.]},
  \bibinfo{author}{Togneri\xfnm[ R.]}.
\newblock \bibinfo{title}{{Cost-Sensitive Learning of Deep Feature
  Representations From Imbalanced Data}}.
\newblock \emph{\bibinfo{journal}{IEEE Transactions on Neural Networks and
  Learning Systems}}
  \bibinfo{year}{2017};\hspace{0pt}:\bibinfo{pages}{1--15}\doi{\bibinfo{doi}{10.1109/TNNLS.2017.2732482}}.
\newblock \eprint{1508.03422}.
\bibitem[{LeCun et~al.(1989)LeCun, Boser, Denker, Henderson, Howard, Hubbard
  et~al.}]{LeCun1989}
\bibinfo{author}{LeCun\xfnm[ Y.]}, \bibinfo{author}{Boser\xfnm[ B.]},
  \bibinfo{author}{Denker\xfnm[ J.S.]}, \bibinfo{author}{Henderson\xfnm[ D.]},
  \bibinfo{author}{Howard\xfnm[ R.E.]}, \bibinfo{author}{Hubbard\xfnm[ W.]},
  et~al.
\newblock \bibinfo{title}{{Backpropagation Applied to Handwritten Zip Code
  Recognition}}.
\newblock \emph{\bibinfo{journal}{Neural Computation}}
  \bibinfo{year}{1989};\hspace{0pt}\textbf{\bibinfo{volume}{1}}(\bibinfo{number}{4}):\bibinfo{pages}{541--551}.
\newblock \doi{\bibinfo{doi}{10.1162/neco.1989.1.4.541}}.
\bibitem[{Yan et~al.(2015)Yan, Chen, Shyu and Chen}]{Yan2015}
\bibinfo{author}{Yan\xfnm[ Y.]}, \bibinfo{author}{Chen\xfnm[ M.]},
  \bibinfo{author}{Shyu\xfnm[ M.l.]}, \bibinfo{author}{Chen\xfnm[ S.c.]}.
\newblock \bibinfo{title}{{Deep Learning for Imbalanced Multimedia Data
  Classification}}.
\newblock In: \emph{\bibinfo{booktitle}{2015 IEEE International Symposium on
  Multimedia (ISM)}}. \bibinfo{publisher}{IEEE}.
\newblock ISBN \bibinfo{isbn}{978-1-5090-0379-2}; \bibinfo{year}{2015},
  \hspace{0pt}p. \bibinfo{pages}{483--488}.
\newblock \doi{\bibinfo{doi}{10.1109/ISM.2015.126}}.
\bibitem[{Wei et~al.(2015)Wei, Gao and Wu}]{Wei2015}
\bibinfo{author}{Wei\xfnm[ X.s.]}, \bibinfo{author}{Gao\xfnm[ B.b.]},
  \bibinfo{author}{Wu\xfnm[ J.]}.
\newblock \bibinfo{title}{{Deep Spatial Pyramid Ensemble for Cultural Event
  Recognition}}.
\newblock In: \emph{\bibinfo{booktitle}{2015 IEEE International Conference on
  Computer Vision Workshop (ICCVW)}}. \bibinfo{publisher}{IEEE}.
\newblock ISBN \bibinfo{isbn}{978-1-4673-9711-7}; \bibinfo{year}{2015},
  \hspace{0pt}p. \bibinfo{pages}{280--286}.
\newblock \doi{\bibinfo{doi}{10.1109/ICCVW.2015.45}}.
\bibitem[{George(2015)}]{George2015}
\bibinfo{author}{George\xfnm[ M.]}.
\newblock \bibinfo{title}{{Image parsing with a wide range of classes and
  scene-level context}}.
\newblock In: \emph{\bibinfo{booktitle}{2015 IEEE Conference on Computer Vision
  and Pattern Recognition (CVPR)}}. \bibinfo{publisher}{IEEE}.
\newblock ISBN \bibinfo{isbn}{978-1-4673-6964-0}; \bibinfo{year}{2015},
  \hspace{0pt}p. \bibinfo{pages}{3622--3630}.
\newblock \doi{\bibinfo{doi}{10.1109/CVPR.2015.7298985}}.
\newblock \eprint{1510.07136}.
\bibitem[{Mostajabi et~al.(2015)Mostajabi, Yadollahpour and
  Shakhnarovich}]{Mostajabi2015}
\bibinfo{author}{Mostajabi\xfnm[ M.]}, \bibinfo{author}{Yadollahpour\xfnm[
  P.]}, \bibinfo{author}{Shakhnarovich\xfnm[ G.]}.
\newblock \bibinfo{title}{{Feedforward semantic segmentation with zoom-out
  features}}.
\newblock In: \emph{\bibinfo{booktitle}{2015 IEEE Conference on Computer Vision
  and Pattern Recognition (CVPR)}}. \bibinfo{publisher}{IEEE}.
\newblock ISBN \bibinfo{isbn}{978-1-4673-6964-0}; \bibinfo{year}{2015},
  \hspace{0pt}p. \bibinfo{pages}{3376--3385}.
\newblock \doi{\bibinfo{doi}{10.1109/CVPR.2015.7298959}}.
\newblock \eprint{1412.0774}.
\bibitem[{Cenggoro(2016)}]{Cenggoro2016a}
\bibinfo{author}{Cenggoro\xfnm[ T.W.]}.
\newblock \emph{\bibinfo{title}{{Imbalanced Learning in Land Use/Land Cover
  Data of Urban Area Using Variational Semi Supervised Learning Embedded with
  Class Expert Layer}}}.
\newblock Master's thesis; Bina Nusantara University; \bibinfo{year}{2016}.
\bibitem[{Goodfellow et~al.(2014)Goodfellow, Pouget-Abadie, Mirza, Xu,
  Warde-Farley, Ozair et~al.}]{Goodfellow2014}
\bibinfo{author}{Goodfellow\xfnm[ I.J.]}, \bibinfo{author}{Pouget-Abadie\xfnm[
  J.]}, \bibinfo{author}{Mirza\xfnm[ M.]}, \bibinfo{author}{Xu\xfnm[ B.]},
  \bibinfo{author}{Warde-Farley\xfnm[ D.]}, \bibinfo{author}{Ozair\xfnm[ S.]},
  et~al.
\newblock \bibinfo{title}{{Generative Adversarial Networks}}.
\newblock \emph{\bibinfo{journal}{Advances in Neural Information Processing
  Systems 27}}
  \bibinfo{year}{2014};\hspace{0pt}:\bibinfo{pages}{2672--2680}\eprint{1406.2661}.
\bibitem[{Liu et~al.(2015)Liu, Luo, Wang and Tang}]{liu2015deep}
\bibinfo{author}{Liu\xfnm[ Z.]}, \bibinfo{author}{Luo\xfnm[ P.]},
  \bibinfo{author}{Wang\xfnm[ X.]}, \bibinfo{author}{Tang\xfnm[ X.]}.
\newblock \bibinfo{title}{{Deep Learning Face Attributes in the Wild}}.
\newblock In: \emph{\bibinfo{booktitle}{2015 IEEE International Conference on
  Computer Vision (ICCV)}}. \bibinfo{publisher}{IEEE}.
\newblock ISBN \bibinfo{isbn}{978-1-4673-8391-2}; \bibinfo{year}{2015},
  \hspace{0pt}p. \bibinfo{pages}{3730--3738}.
\newblock \doi{\bibinfo{doi}{10.1109/ICCV.2015.425}}.
\bibitem[{Ioffe and Szegedy(2015)}]{Ioffe2015}
\bibinfo{author}{Ioffe\xfnm[ S.]}, \bibinfo{author}{Szegedy\xfnm[ C.]}.
\newblock \bibinfo{title}{{Batch Normalization: Accelerating Deep Network
  Training by Reducing Internal Covariate Shift}}.
\newblock \emph{\bibinfo{journal}{Proceedings of The 32nd International
  Conference on Machine Learning}}
  \bibinfo{year}{2015};\hspace{0pt}\eprint{1502.03167}.
\bibitem[{Krizhevsky and Hinton(2012)}]{Krizhevsky2012}
\bibinfo{author}{Krizhevsky\xfnm[ A.]}, \bibinfo{author}{Hinton\xfnm[ G.E.]}.
\newblock \bibinfo{title}{{ImageNet Classification with Deep Convolutional
  Neural Networks}}.
\newblock In: \emph{\bibinfo{booktitle}{Advances in Neural Information
  Processing Systems}}. \bibinfo{year}{2012}, \hspace{0pt}.
\bibitem[{Kingma and Ba(2015)}]{Kingma2015}
\bibinfo{author}{Kingma\xfnm[ D.P.]}, \bibinfo{author}{Ba\xfnm[ J.]}.
\newblock \bibinfo{title}{{Adam: A Method for Stochastic Optimization}}.
\newblock \emph{\bibinfo{journal}{The International Conference on Learning
  Representations 2015}}
  \bibinfo{year}{2015};\hspace{0pt}:\bibinfo{pages}{1--15}\eprint{1412.6980}.

\end{thebibliography}
\clearpage

\end{document}